\documentclass[letterpaper]{article}
\usepackage{aaai20}
\usepackage{times}
\usepackage{helvet}
\usepackage{courier}
\usepackage[hyphens]{url}
\usepackage{graphicx}
\usepackage{graphicx}
\title{One Homonym per Translation}
\author{Bradley Hauer \and Grzegorz Kondrak\\
Department of Computing Science\\
University of Alberta, Edmonton, Canada\\
{\tt {\{bmhauer,gkondrak\}}@ualberta.ca}
}

\newcommand{\newcite}[1]{\cite{#1}}

\usepackage{amsmath}
\usepackage{multirow}

\newcommand{\bhap}{between homonymy and polysemy}
\newcommand{\etymo}{Type-A}
\newcommand{\nonetymo}{Type-B}

\newcommand{\SI}{sense inventory}

\newcommand{\hcs}{homonym resource}
\newcommand{\nsix}{ODE clustering}

\begin{document}
\maketitle
\begin{abstract}
The study of homonymy is vital to resolving
fundamental problems in lexical semantics.
In this paper,
we propose
four hypotheses
that characterize the unique behavior of homonyms
in the context of translations, discourses, collocations, and sense clusters.
We present a new annotated {\hcs} 
that allows us to test our hypotheses on existing WSD resources. 
The results of the experiments 
provide strong empirical evidence for the hypotheses.
This study represents a step towards
a computational method for distinguishing between homonymy and polysemy, 
and constructing a definitive inventory of coarse-grained senses.
\end{abstract}

\section{Introduction}
\label{intro}

Many words are semantically ambiguous, 
in that 
they have multiple senses.
The relationship between two senses of a
word is called {\em polysemy} if they are semantically related, 
and {\em homonymy} otherwise \cite{jm2}.
Senses that belong to the same homonym are polysemous 
(e.g. \#2 and \#5 in Table~\ref{bank_senses}),
while senses of distinct homonyms are homonymous
(e.g. \#2 and \#1 in Table~\ref{bank_senses}).

The differentiation of homonymous and polysemous word senses is 
one of the central problems
of lexicography
\cite{melcuk}.
A textbook on thoeretical semantics 
devotes an entire chapter to the problem,
concluding that it may be insoluble,
as the intuitions of native speakers cannot be relied upon \cite{lyons1995}.
Psycho-linguistics furnishes evidence 
for a common representation of closely related senses in the mental lexicon
\cite{brown2008},
which suggests that NLP applications would benefit from the ability
to distinguish homonym-level meaning differences 
\cite{utt2011}.
In fact,
standard NMT systems make a substantial number of errors on homonyms \cite{liu2018}.

The study of homonymy is also 
of utmost importance to the problem of 
establishing the set of senses for a given word.
In word sense disambiguation (WSD),
which is the task of selecting the intended sense of an ambiguous word token,
the quality and granularity of the {\SI}
greatly influences the design, evaluation, and utility of any system. 
The standard {\SI}, WordNet \cite{fellbaum1998},
makes no distinction {\bhap}, and
is widely considered to be excessively fine-grained
for many practical applications 
\cite{navigli2018},
as evidenced by a low inter-annotator agreement
\cite{snyder2004}.
This has inspired substantial prior work on clustering fine-grained senses
to create more coarse-grained sense inventories
\cite{hovy2006,navigli2006,snow2007,dandala2013,mccarthy2016}.

\begin{table}
 \centering
 \small
  \begin{tabular}{|ll|ll|}
\multicolumn{2}{c}{BANK$_n^1$} & \multicolumn{2}{c}{BANK$_n^2$} \\ \hline
    \#2  & financial institution & \#1  & sloping land \\
    \#5  & stock held in reserve & \#3  & long ridge or pile \\
    \#6  & funds held by a house \;\; & \#4  & arrangement of objects \\
    \#8  & container for money & \#7  & slope in
a road \\
    \#9  & building  & \#10 & flight maneuver \\
    \hline
  \end{tabular}
  \caption{The senses of the noun `bank' from WordNet 3.0,
           grouped by its two homonyms.}
  \label{bank_senses}
\end{table}

Following the observation that
different senses of a word often correspond to distinct words in another language
\cite{resnikyarowsky1997}, 
another branch of prior work
has sought to use translations to define sense inventories
\cite{resnik1999,diab2002,ng2003,chan2007,apidianaki2008,bansal2012,taghipour2015}.
In order to be successful,
such an approach would have to resolve the challenging issues
of mapping senses to translations in a set of diverse target languages,
as well as projecting them onto a standard sense inventory, such as WordNet.

In summary,
clustering fine-grained senses and defining sense distinctions using translations 
are two competing methodologies for creating coarse-grained sense inventories.
Regardless of which 
one is adopted,
an understanding of the nature and characteristics of homonymous senses
is a necessary step toward a principled method
of defining senses and sense distinctions.
In particular, distinctions between homonymous senses must be
preserved in any sense inventory.
This motivates our study,
which contributes to such an understanding
by directly linking homonymy to the concepts of translation and sense clustering,
and thus bridging the gap between the two approaches.

The contributions of this work are both theoretical and empirical.
The main goal is to create theoretical foundations for the study of homonymy,
which could pave the way for 
developing a computational method for distinguishing {\bhap},
and facilitate the task of constructing
a definitive inventory of coarse-grained senses.
We propose four hypotheses about the unique behavior of homonyms 
in the context of translations, discourses, collocations, and sense clusters.
The hypotheses are formulated using established semantic concepts,
and formalized in mathematical notation.
Our principal hypothesis, as stated in the title,
implies a sufficient condition for polysemy
which is
observable and replicable.

Apart from introducing the hypotheses,
we perform experiments to provide empirical evidence for them.
It is clear from prior work that what is true at one level of semantic granularity
may not be true at another.
For example,
the well-known hypotheses 
\emph{one sense per discourse} 
and \emph{one sense per collocation}
have been found not to hold consistently for WordNet senses.
It is critical
that all claims be formally stated and experimentally tested,
regardless of whether
the results are considered surprising;
we have found no prior work that 
fulfills this requirement with respect to
the four hypotheses presented in this paper.
To facilitate our experiments,
we create a new annotated resource,
by identifying nearly three thousand English homonyms,
and mapping them onto WordNet senses.
The results of our experiments 
on multiple annotated corpora and language pairs
strongly support our hypotheses.

\section{Homonym Hypotheses}
\label{hypotheses}

In this section, 
we formally define the notion of a homonym,
and formulate our hypotheses using set notation.
We attempt to keep the notational complexity to a minimum,
while at the same time striving to avoid ambiguity.

\subsection{Preliminaries}
\label{prelim}

{\em Lexemes} are 
units of language that are represented in the lexicon \cite{murphy10}.
{\em Words} are sets of word-forms that represent lexemes,
and are associated with certain morpho-syntactic properties.
This definition of words includes compounds, such as `single out', 
as is the case in WordNet.
We consider both lexemes and words
that differ in part of speech as distinct.
We write lexemes in capital letters, abstract words in single quotes,
actual word-forms in italics, and sense meanings in double quotes.
For example, 
the lexeme CUT$_{v}$ is represented by 
the verb `cut', with 
the word-forms {\em cut, cuts,} and {\em cutting}.
A lexeme is called polysemous if it contains multiple senses,
and monosemous if it has only a single sense.
Senses that belong to the same lexeme are semantically related,
and therefore 
polysemous
\cite{jm2}.

A \emph{homonymous word} (e.g., the noun `bank' in Table~\ref{bank_senses})
represents more than one lexeme,
and those lexemes 
are called \emph{homonyms}.
Senses
associated with distinct homonyms 
are unrelated
and therefore
homonymous
\cite{murphy10}.
Consequently, the problem of deciding whether two senses of a homonymous word
are polysemous
is equivalent
to deciding whether
they belong to the same lexeme.
Furthermore, since a non-homonymous word represents only a single lexeme,
all of its senses are polysemous.

We are now ready to formally define homonyms.
Let $\mathcal{L}$ and $\mathcal{W}$ denote the sets of lexemes
and words of a given language, respectively,
and let $w$: $\mathcal{L} \mapsto \mathcal{W}$
be a function that maps each lexeme to the word that represents it.
In later sections, we will use 
$w^{-1}$: $\mathcal{W} \mapsto \mathcal{P}(\mathcal{L})$ 
to denote the function which maps each word 
to the set of lexemes it represents.
We define the set of homonymous words $\mathcal{H}$ as
the set of all words that represent multiple lexemes: 

\medskip
$ \indent\indent
  \mathcal{H} \overset{\underset{\mathrm{def}}{}}{=}
  \{
  W \in \mathcal{W} \;|\;
  \exists L, L' \in \mathcal{L}:
  \\\indent\indent\indent\indent
  (L \neq L') \,\land\, (w(L) = w(L') = W)
  \}$\\

For example,
$w$(\text{BANK$_n^1$}) = $w$(\text{BANK$_n^2$}) = `bank' $\in \mathcal{H}$.

\subsection{One Homonym per Translation}
\label{ohpt}

In general,
there is no simple correspondence between word senses
and their translations:
a single sense may be translated by any of several synonyms,
and 
different senses of the same word 
may have the same translation.
\newcite{ide2007} observe that
cross-lingual distinctions
often correspond to homonym-level disambiguation.
We posit
a direct relationship between translations and homonyms.
Intuitively,
if we randomly selected two different words from a bilingual dictionary,
we would not expect them to have translations in common.
The same reasoning applies to homonyms,
since
they are semantically unrelated lexemes that coincidentally share the same form.
We formalize this insight as 
our principal hypothesis.

Put simply,
the one homonym per translation hypothesis (OHPT)  
states that
homonyms
have disjoint translation sets.
Formally, 
let $T(L)$ be a set of translations of a lexeme $L$,
and let $w^{-1}$
be as defined as in Section \ref{prelim}.
Then, 

\bigskip
$ \indent\indent
  \forall H \in \mathcal{H}: \forall L,L' \in w^{-1}(H): 
  \\\indent\indent\indent\indent
  (L \neq L') \Rightarrow T(L) \cap T(L') = \emptyset $\\

For example, 
the Italian translations of the noun `yard' can be partitioned into two
disjoint sets
$T(\mbox{YARD}^1_{n})$  = \{`iarda',`yard'\} and
$T(\mbox{YARD}^2_{n})$  = \{`cortile',`giardino'\},
which correspond to two English homonyms,
with the meanings of ``unit'' and ``garden'', respectively.

This hypothesis 
implies an important generalization:
{\em the existence of a shared translation 
is a sufficient condition for polysemy}.
Indeed,
for homonymous words,
senses that can be translated by the same word must belong to the same lexeme,
and so are polysemous.
As all other words represent only single lexemes,
all their senses are polysemous by definition
(Section~\ref{prelim}).  
Therefore, 
we consider the OHPT hypothesis as a
major step towards solving the problem of distinguishing {\bhap}.

\subsection{One Homonym per Discourse}
\label{ohpd}

The \emph{one sense per discourse} (OSPD) hypothesis
was introduced in the seminal paper of \newcite{gale1992}.
They observe that 
``well-written discourses tend to avoid multiple senses of a polysemous word",
and confirm that the property holds with high probability on a set of 82
instance pairs involving 9 ambiguous words.
However, \newcite{krovetz1998}
reports that OSPD 
holds for only 67\% of ambiguous words in SemCor, 
and conjectures that 
the hypothesis may only apply to homonymous senses. 
 
We formulate Krovetz's conjecture as
the \emph{one homonym per discourse} hypothesis (OHPD),
which can be viewed as a specialization of OSPD to homonyms.
The hypothesis states that 
{\em all occurrences of a homonymous word in
a discourse represent the same homonym}.
A possible explanation of this phenomenon is that
writers avoid the use of homonyms by employing their synonyms
in order to reduce ambiguity in a discourse.
Another explanation is that most discourses
cover topics within a single domain,
and therefore are unlikely to contain lexemes that are 
completely unrelated to each other.

Our formulation of the OHPD hypothesis
states that no more than one lexeme of a homonymous word
occurs in any given discourse.
Formally, 
let $D$ be the set of lexemes that occur in a discourse,
and 
let $w$ be again the function that maps lexemes to words.
Then,
\[ \forall L,L' \in D: (w(L) = w(L')) \Rightarrow (L = L') \]

\begin{figure}[t]
  \centering
  \includegraphics[scale=0.5]{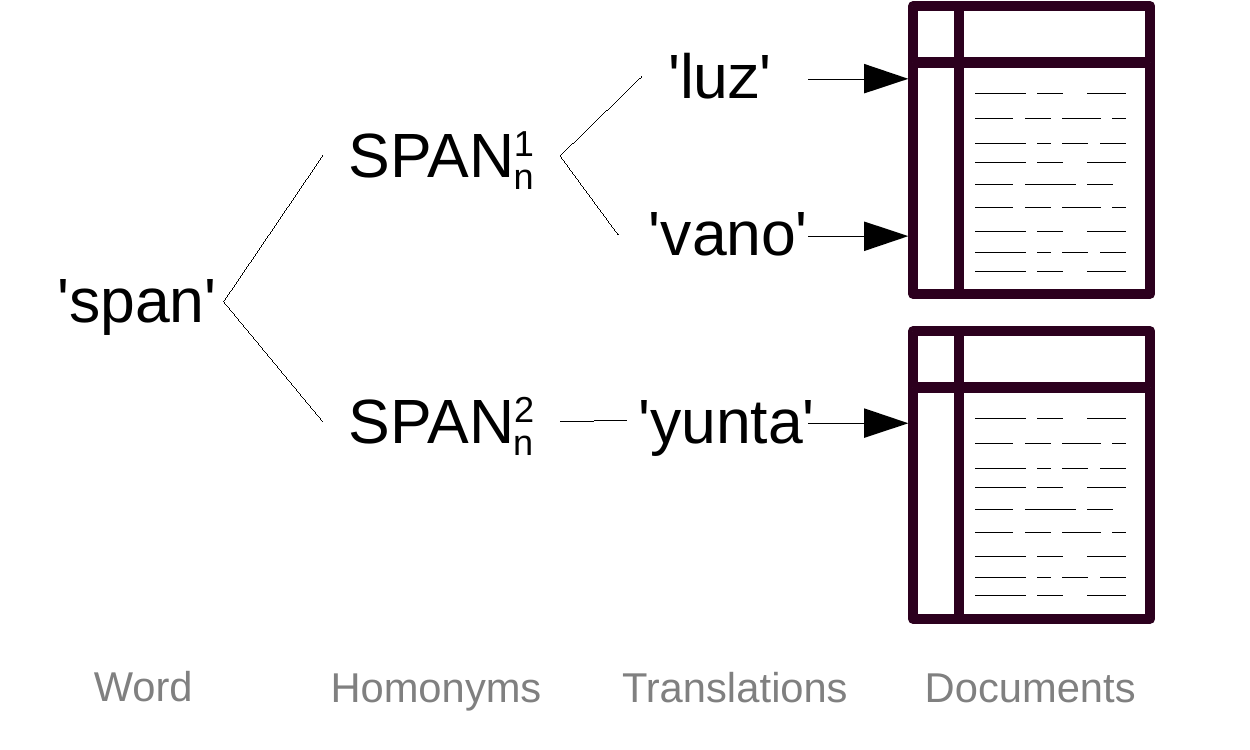}
  \caption{An example of an exception to the {\em one translation per discourse}
hypothesis of \newcite{carpuat2009}.
   The top two Spanish translations of 'span' are synonymous.}
  \label{whtd}
\end{figure}

We close this section by considering the relationship between OHPD and
the {\em one translation per discourse} (OTPD) hypothesis
of \newcite{carpuat2009}.
They report that 
approximately 80\% of French words 
have a single English translation per document,
which they interpret as strong support for their hypothesis. 
We note that 
the conjunction of our OHPT and OHPD hypotheses does not imply OTPD.
Indeed,
consider the example
in Figure~\ref{whtd},
which shows how the occurrence of three Spanish translations 
of the homonymous noun `span' in two different documents
leads to a violation of OTPD, but not of OHPD or OHPT.

\subsection{One Homonym per Collocation}
\label{ohpc}

\newcite{yarowsky1993} proposes
the \emph{one sense per collocation} (OSPC) hypothesis,
broadly defining a collocation as
``the co-occurrence of two words in some defined relationship''.
Yarowsky reports that the hypothesis holds 
with the average 95\% precision 
on a sample of words of an unreported size.
However,
\newcite{martinez2000} 
find much weaker evidence for OSPC on WordNet senses,
with precision values rarely exceeding 70\%.

The explicit focus of \newcite{yarowsky1993} 
is on the most coarse-grained sense distinctions.
Their word sample
includes 
pseudo-words,
words with different French translations,
words spelled the same but pronounced differently ({\em homographs}), 
words pronounced the same but spelled differently ({\em homophones}), 
and words that are visually confusable in optical character recognition. 
All these types of words can be viewed as approximations of homonymy,
as they involve
pairs of distinct lexemes.
We formalize this notion with
the \emph{one homonym per collocation} (OHPC) hypothesis,
which states that 
only one homonym of a word should appear in any given collocation.

Formally,
given a corpus of text,
let $\mathcal{R}$ be the set of all
collocations.
For lexeme $L \in \mathcal{L}$,
and collocation $r \in \mathcal{R}$,
let $C_r(L)$ be a proposition which is true
if and only if $w(L)$ occurs in collocation $r$ in the corpus.
Then,

\bigskip
$
  \indent
  \forall H \in \mathcal{H}: 
  \forall L,L' \in w^{-1}(H):
  \forall r \in \mathcal{R}:
  \\\indent\indent\indent\indent
  (C_r(L) \land C_r(L'))
  \Rightarrow 
  (L = L') 
$\\

For example, 
if BANK$_n^1$ (``repository'')
is found to occur in the collocation [word-to-right = {\em hired}]
then 
BANK$_n^2$ (``ridge'') is unlikely to occur in this collocation.

\subsection{One Homonym per Sense Cluster}
\label{ohpsc}

\emph{Sense clustering} is the task of
grouping together senses that are closely related \cite{dandala2013}.
Although
the criteria for
eliminating
sense distinctions
vary depending on the purpose of the sense inventory,
a common motivation is to reduce
the excessive granularity of WordNet \cite{snow2007}.
In particular,
a manual clustering of WordNet senses was created
as part of the OntoNotes project, 
with the objective of increasing the inter-annotator agreement on WSD to 90\%
\cite{hovy2006}.
Sense clustering has been shown to improve performance on a number of NLP tasks
\cite{pilehvar2017},
and can serve as an extrinsic evaluation 
for learned representations of senses \cite{mancini2017}.

Since homonyms are distinct lexemes,
we posit that any well-grounded clustering approach
must avoid merging homonymous senses.
Formally,
let $\mathcal{C}$ be a sense clustering, a set of disjoint sets of senses,
and let $S(L)$ be the set of senses of lexeme $L$.
Then,
\[
 \forall C \in \mathcal{C}: 
 \exists L \in \mathcal{L}: C \subseteq S(L)
\]

In plain words, 
while the senses of a homonym may be divided between multiple clusters,
no cluster should contain senses from different homonyms.

\section{Homonym Data}
\label{homonym_data}

In order to provide experimental evidence for our homonym hypotheses,
we need a large set of ``gold'' homonyms,
as well as a mapping between those homonyms 
and the sense annotations in existing corpora.
Since no such resource is publicly available,
we create our own collection of
English homonyms
(see Table~\ref{homonym_table}).
In this section, 
we present a binary typology of homonyms,
our methodology for creating a list of homonyms,
and
the method for mapping those homonyms onto the WordNet sense inventory.

\subsection{Typology of Homonyms}
\label{typology}

There are generally two ways of defining homonyms.
In linguistics
(and in this paper),
homonyms are considered to be distinct lexemes 
that happen to share the same form
\cite{murphy10}.
In lexicography,
homonymy is sometimes defined more narrowly,
by additionally requiring the etymological origins of the lexemes to be different
\cite{ode}.
Homonyms can therefore be divided into two types:
those that satisfy the requirement of different origins,
and those that do not.
Due to the lack of commonly-accepted terminology,
we refer to these two types of homonyms simply as
{\etymo} and {\nonetymo}, respectively.

\begin{figure}[t]
  \includegraphics[width=\linewidth]{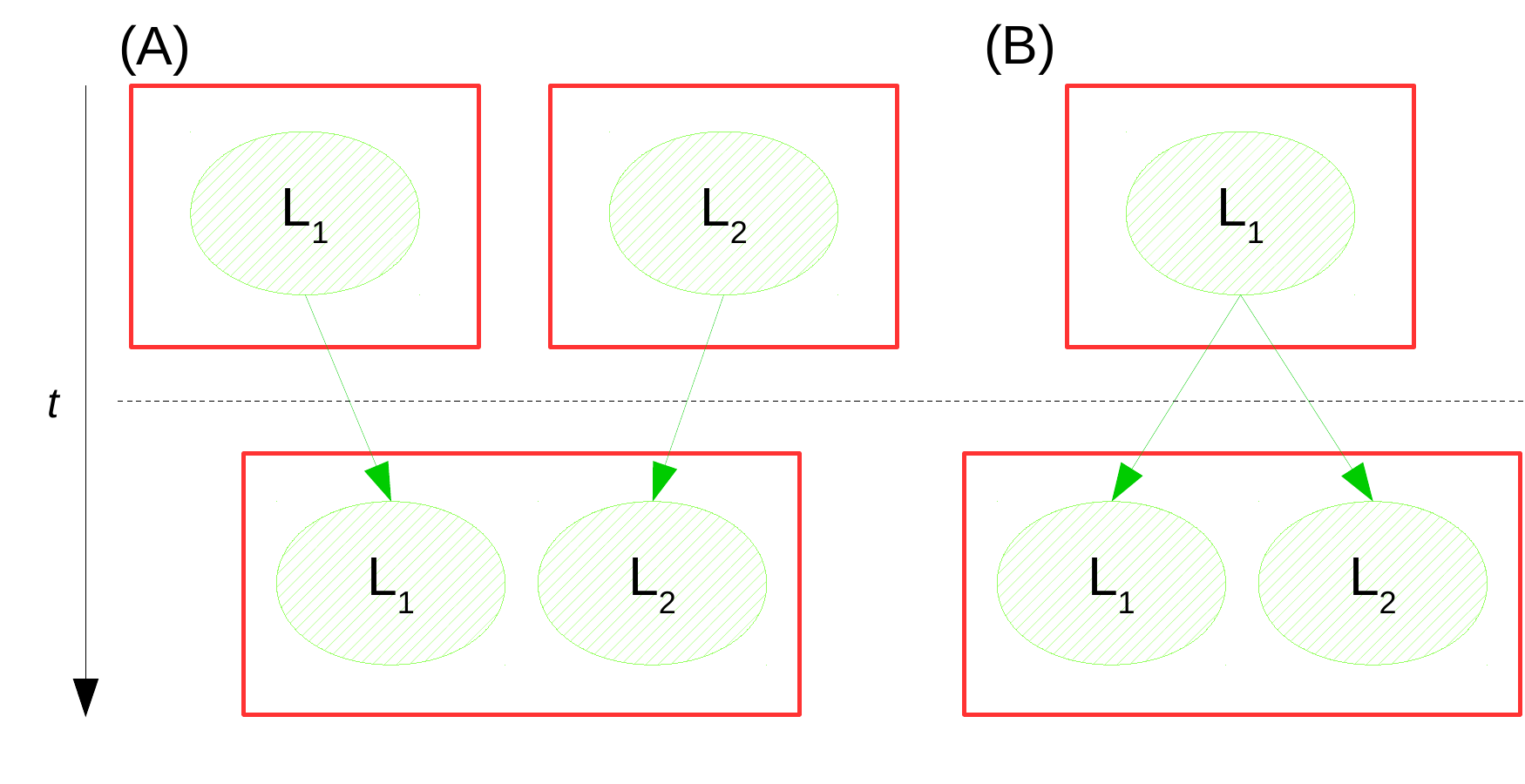}
  \caption{A schematic illustration of the diachronic distinction between
  two types of homonyms.
  Circles represent lexemes; boxes represent words.
  }
  \label{boxes}
\end{figure}

The two types of homonyms,
which are schematically illustrated in Figure~\ref{boxes},
stem from different diachronic phenomena.
{\etymo} homonyms arise from 
a convergence of distinct words
into a single form.
This can occur through the process of sound change 
or inter-lingual borrowing.
For example, 
both the Old English word {\em c{\ae}g} ``locking implement''
and the 17th-century Spanish borrowing {\em cayo} ``island''
evolved into the modern English 
{\em key}.
{\nonetymo} homonyms, on the other hand,
arise when a single lexeme splits into two lexemes
due to the process of semantic drift.
For example, 
the two meanings of 
{\em staff},  ``pole'' and ``people'', have developed 
from a single etymon, 
which is attested in Old English as {\em st{\ae}f}.
Importantly,
as native speakers are generally unaware of
the etymological history of words,
these two types of homonyms are indistinguishable 
in the synchronic analysis of languages
\cite{lyons1995}. 

The crucial methodological advantage of {\etymo} homonyms is that
they can be objectively identified
by consulting existing etymological dictionaries.
Even though the process of
compiling an exhaustive list of {\etymo} homonyms 
for any language is time-consuming,
it is still much easier and less controversial than 
conducting psychological experiments with human subjects
\newcite{brown2008}, or
obtaining consensus within teams of linguistic experts
\newcite{ontonotes}.
We have accomplished this task for English 
by creating a {\hcs} 
that we describe next.

\subsection{List of {\etymo} Homonyms}
\label{resource}

The new {\hcs}\footnote{\it{https://webdocs.cs.ualberta.ca/\~{}kondrak}}, which
enables us to empirically test our homonym hypotheses,
contains words that represent multiple
lexemes with distinct etymological origins.
We compiled the list by collecting all homonyms 
that we could find in dictionaries,
including the English Oxford Living 
Dictionary\footnote{\it{https://en.oxforddictionaries.com}}
and the Concise Oxford Dictionary of English 
Etymology\footnote{\it{http://www.oxfordreference.com}}.
We include 
all homonyms that
at some
point during 
language evolution
existed as
separate words,
even those
that can be traced to a single proto-word.
For example, 
we include the homonyms of the noun {\em sole}
(``undersurface'' vs.\@ ``fish'')
because of their distinct histories,
even though both ultimately come from
Latin {\em solea} ``sandal''.

\begin{table}[t]
  \centering
  \begin{tabular}{ccccccc}
    \hline
    POS & Origin & Gloss & French \\
    \hline
    N,V & Old French \emph{espan}
      & distance & \emph{port\'{e}e} \\
    N,V & Low German \emph{spannen}
      & rope     & \emph{filin}      \\
    Adj   & Old Norse \emph{sp\'{a}n-n\'{y}r}
      & clean    & \emph{impeccable} \\
    V   & Old English \emph{spinnan}
      & rotate   & \emph{tourn\'{e}} \\ \hline
  \end{tabular}
  \caption{Sample entries of the \hcs{}, which correspond to
six homonyms of the English lemma {\em span}.}
  \label{homonym_table}
\end{table}

Table~\ref{homonym_table} shows sample entries from our resource.
The list contains 2759 {\etymo} homonyms that correpond to 
804 lemmas,
1601 unique lemma/POS pairs,
and 1967 distinct etymologies. 
The number of distinct etymologies per lemma ranges from two to six.
Each entry includes 
etymological information
(the form and the language of origin),
and a list of possible parts of speech (noun, verb, adjective, adverb).
For the purpose of disambiguation
in subsequent stages of annotation,
each entry was manually assigned
a brief English gloss,
as well as a single French translation.
We excluded from our list all proper nouns and abbreviations.

About two dozen of the homonymous words in our resource represent homographs,
which are homonyms that differ in pronunciation.
For example, the noun `bass' 
is pronounced [b{\ae}{s}] or [bes]
depending on whether it refers to 
a fish or a musical instrument,
respectively.  
Although most of the dictionary words with alternative pronunciations appear
to involve {\etymo} homonyms,
we found a number of exceptions.
They include 
{\nonetymo} homonyms (e.g. `pension'),
polysemous words (e.g. `undertaking'),
common vs.\@ proper nouns (e.g. `job'),
matching word-forms of distinct lemmas (e.g. `putter'),
as well as pronunciation variants (e.g. `puissance').
Since our focus is on written language, 
our resource excludes homophones,
such as `cellar' vs. `seller'.

Although we make no claim about the completeness of our {\hcs},
we consider it to be representative of English homonyms in general.
This is based on the fact that {\etymo} and {\nonetymo} homonyms
cannot be distinguished without access to etymological expertise.

\subsection{Mapping WordNet Senses to Homonyms}
\label{mapping}

In order to test our homonym hypotheses,
we must be able to convert the existing word sense annotations 
into homonym annotations.
For example,
we need to know 
which homonym from our list is represented by
a word token {\em spans} 
which is sense-annotated 
as 
``two items of the same kind''
in some corpus.
The standard sense inventory for WSD is WordNet.
In this section,
we describe our method of mapping  
the homonyms in our new resource to
WordNet senses.

Because of the large number of fine-grained senses in WordNet,
it was not practical to directly map each WordNet sense
of each homonymous word to the corresponding homonym.
Instead, we made use of 
the existing clustering \cite{navigli2006},
which 
was created by automatically mapping
WordNet 2.1 senses to more coarse-grained senses 
defined by the Oxford Dictionary of English (ODE).
Our 804 homonymous lemmas correspond to 
2644 sense clusters, which contain 5361 senses.
We manually mapped each cluster of senses to a single homonym
on the basis of their WordNet sense glosses.

The resulting mapping is imperfect 
for two reasons.
First, the {\nsix} itself
is not always correct,
which sometimes results
in homonymous senses being placed in the same cluster.
Second, our human annotator
made some errors in mapping clusters to homonyms.
We performed the following validation experiment in order to
estimate the accuracy of the overall mapping.
A second annotator performed a direct 
mapping of 268 WordNet senses corresponding to 
a random sample of 77 homonymous words,
without any reference to the {\nsix}.
We found that the two independent mappings of the 268 senses
differed in only 17 instances,
which implies that the overall error rate has an upper bound of 6\%.

The errors in the sense-to-homonym mapping 
are a source of ``false alarms'' 
in the experiments described in Section~\ref{exp}.
We are confident in our ability to 
determine which of the apparent exceptions are actual exceptions
to our hypotheses
by careful analysis of the available data.
While the distinction {\bhap} can be highly subjective,
the mapping of WordNet senses to known homonyms is much easier,
as confirmed by our validation experiment described above.

\section{Homonym Evidence}
\label{exp}

In this section,
we describe the experiments that test
the four hypotheses formulated in Section~\ref{hypotheses}
using 
the full set of homonyms in 
our new {\hcs} from
Section~\ref{homonym_data}.

\subsection{SemCor and Translations}
\label{semcor}

For testing the OHPD
and OHPC
hypotheses,
we use SemCor \cite{miller1993},
a large sense-annotated English corpus 
which was created as part of the WordNet project \cite{petrolito2014}.
In particular,
we adapt the version of SemCor from
\newcite{raganato2017}.\footnote{\it{http://lcl.uniroma1.it/wsdeval}}
The number of word tokens, types, and senses
are in Table~\ref{table_msc}
(words are defined as lemma/POS pairs)

For testing the OHPT hypothesis,
we require not only sense annotations,
but also the corresponding translations.
At the minimum,
we need a large word-aligned bitext that
has both sense and part-of-speech annotations on
the source side,
and lemma annotations on both sides.
In addition, the sense inventory has to 
be the same as the one in our {\hcs}.
Although such resources are rare,
we managed to adapt two bitexts
to meet these requirements:
MultiSemCor \cite{bentivogli2005},
and JSemCor \cite{bond2012}.
These corpora,
which we refer to as MSC and JSC,
contain partial word-aligned
translations of SemCor into Italian and Japanese, respectively.

\subsection{WordNet}
\label{data}

The use of WordNet presents a number of technical challenges.
For the purpose of replicability,
we describe here two major     issues.

The first issue concerns
two distinct conventions
for referring to individual WordNet senses:
{\em sense keys} (used in SemCor, JSC, and the \nsix{})
and {\em sense numbers} (used in MSC and OntoNotes).
We converted the former into the latter
using the WordNet::SenseKey 
package.\footnote{\it{https://metacpan.org/release/LINAS/WordNet-SenseKey-1.03}}
Because the mapping is not always one-to-one,
16 out of 60,655 WordNet senses in the \nsix{}
had to be excluded;
however, none of the affected words are in our \hcs{}. 

The second issue is the mapping between different WordNet versions.
We converted the sense keys from  WordNet 2.1
-- the version of WordNet used 
in the clustering described in \newcite{navigli2006} --
to
WordNet 3.0
-- the version used by all other resources in this paper -- 
using WordNetMapper.\footnote{\it{https://github.com/cltl/WordNetMapper}}
The package failed to map
551 out of 60,655 senses in the \nsix{},
which resulted in 22 WordNet senses being excluded from our \hcs{}.
Due to these issues, we decided not to further map all WordNet senses
in our resources to WordNet 3.1.

\begin{table}[t]
\centering
\begin{tabular}{cccc}
\hline
\multicolumn{1}{c}{}
& \multicolumn{1}{c}{SemCor}
& \multicolumn{1}{c}{MSC}
& \multicolumn{1}{c}{JSC} \\\hline
Word tokens    & 226,034 & 92,992 & 58,257  \\
Word types     &  20,399 & 11,451 &  8,445  \\
WordNet senses &  33,308 & 17,875 & 12,516  \\ \hline
\end{tabular}
\caption{The size of the English side of each corpus.}
\label{table_msc}
\end{table}

\subsection{One Homonym per Translation}
\label{ohpt_exp}

\begin{table*}[t]
\centering
\begin{tabular}{cccccccc}
\hline
\multirow{2}{*}{\#} &
\multirow{2}{*}{Hypothesis} &
\multirow{2}{*}{Focus} &
\multirow{2}{*}{Corpus} &
\multirow{2}{*}{Instances} &
\multicolumn{2}{c}{Exceptions} &
\multicolumn{1}{c}{Support} \\
& & & & & apparent & actual & (in \%)\\
\hline
1 & OHPT & translations & MSC (Italian)  & 1093 &  7 &  1 & 99.9 \\
2 & OHPT & translations & JSC (Japanese) & 1093 &  3 &  2 & 99.8 \\
3 & OHPD & documents & SemCor & 2126 & 14 &  9 & 99.6 \\
4 & OHPC & collocations & SemCor &  522 & 16 & 11 & \;\,97.9\footnotemark{}
\\
5 & OHPSC & sense clusters  & OntoNotes & 1578 & 23 & 2 & 99.9 \\
\hline
\end{tabular}
\caption{Summary of the evidence for the homonym hypotheses
from our five experiments.}
\label{results}
\end{table*}

The OHPT hypothesis
characterizes the relationship between homonymous words
and their translations in another language.
We validate the hypothesis on 
two language pairs
using the annotated bitexts described in Section~\ref{semcor}.

In the experimental evaluation,
we compute the percentage of type-level
instances that are consistent with the OHPT hypothesis.
For each English word (i.e. lemma/POS pair)
that appears in our {\hcs},
we identify the set of its translations on the target side of the bitext.
Each unique word/translation pair constitutes a single instance.
An instance is consistent with the OHPT hypothesis
if and only if
all of its occurrences in the bitext
represent the same homonym.
For example, 
the Italian translation `gioco' 
corresponds to three different senses of the noun `game' in MSC,
but since all of them belong to the same homonym,
this instance is consistent with OHPT. 

The results of the evaluation on the MSC and JSC bitexts
are shown in Rows 1 and 2 of Table~\ref{results}.
Coincidentally, 
MSC and JSC have the same number of unique word/translation pairs (1093).
The two corpora contain only 3 actual exceptions to OHPT.
The single actual exception in MSC 
involves the homonyms represented by the noun `band'
which is often translated in Italian as `banda'.
In this
case, the homonymy in English (``ring'' vs.\@ ``group'')
is mirrored by an analogous case of homonymy in Italian.
The two actual exceptions in JSC 
involve the English lexical loans 
`case' and `club',
which
have the same Katakana written form 
regardless of the homonym they represent.
We attribute these exceptions to the phenomenon of {\em parallel homonymy},
which may arise
in the process of lexical borrowing.
 
In addition to the 3 actual exceptions,
the experiment identified 7
exceptions
that are caused by data errors in the two corpora. 
The
data errors
can be divided into four categories:
(1)~
incorrect sense annotations in SemCor,
e.g.  {\em ``the \underline{case} of Jupiter''} 
annotated with the sense of ``container'';
(2)~
an incorrect sense translation in MSC:
{\em flag} in the sense of ``flower''
translated as {\em bandiera} instead of {\em iride};
(3)~
errors in the {\nsix}, 
e.g. two homonymous senses of `club' 
(``team'' and ``playing card'')
in the same cluster;
(4)~
an error in our manual mapping between the {\nsix} and the homonyms:
`light' in the sense of ``free from troubles'' 
being mapped to the homonym ``not dark''.
We conclude that
the OHPT hypothesis is supported in over 99.8\% of instances in either bitext.

In order to verify that 
partitioning of translations is 
a property of homonyms,
and not simply of any sense clusters,
we perform an additional experiment on MSC.
We randomly select two sets of 20 words (i.e. lemma/POS pairs)
from our {\hcs} and the OntoNotes clusters, respectively.
We consider only words 
that are represented in MSC by senses 
from exactly two homonyms or two OntoNotes sense clusters.
None of the OntoNotes words occur in our {\hcs}.
This yields 40 words with
a similar number of sense-annotated tokens:
$6.80$ per homonym,
and $7.25$ per OntoNotes cluster, on average.
We find that 
16 of the 20 homonym pairs,
and 6 of the 20 OntoNotes cluster pairs
exhibit strict translation partitioning in MSC.
In total,
there are 4 instances of overlapping translations between 4 homonym pairs
(a subset of the 7 apparent exceptions in Table~\ref{results}),
and 17 such instances between 14 OntoNotes cluster pairs
(3 cluster pairs share multiple translations).
This result is statistically significant ($p < 0.005$)
according to the $\chi^2$ test.
We conclude that homonyms are significantly more likely 
to exhibit translation partitioning than OntoNotes sense clusters.

\subsection{One Homonym per Discourse}
\label{ohpd_exp}

The OHPD hypothesis predicts that all tokens
of a given homonymous word
in a discourse correspond to the same homonym.
We validate the hypothesis
on English SemCor (Section \ref{semcor}),
taking each of its documents as a single discourse.

In the experimental evaluation, 
we compute the percentage of type-level instances 
that are consistent with the OHPD hypothesis. 
For each English word (i.e. lemma/POS pair) that appears in our {\hcs}, 
we identify all its occurrences in the corpus.
Each unique word/document pair constitutes a single instance.
An instance is consistent with the OHPD hypothesis
if and only if 
all of the occurrences of the word in the document
represent the same homonym. 

When a homonymous word occurs only once in a document,
there is of course no possibility of an actual OHPD violation.
However, we consider those instances to support the hypothesis as well,
because the writer may have chosen 
to replace a homonym with one of its synonyms
in order to avoid potential ambiguity.

The results of the evaluation are shown in Row 3 of Table~\ref{results}.
SemCor is divided into 352 documents,
with an average of 642 sense-annotated open-class words per document.
A careful analysis of the 14 apparent exceptions
reveals that four of them are caused by sense annotation errors in SemCor
(e.g.,
{\em sharp \underline{bow} of a skiff}
 is annotated as ``weapon for shooting arrows''),
and one results from an error in the {\nsix}.
The 9 actual exceptions involve the homonymous nouns
`bank', `lead', `list', `port', `rest', and `yard',
as well as the verb `lie'. 
We conclude that 
fewer than 0.5\% of instances in SemCor contradict the OHPD hypothesis.

\footnotetext{This number is a lower bound estimate.}

\subsection{One Homonym per Collocation}
\label{ohpc_exp}

The OHPC hypothesis predicts that only one homonym of a word appears in any
given collocation.
Due to the broad definition, wide variety, and large number
of possible collocations,
it is difficult to definitively establish the extent to which 
the OHPC hypothesis holds for a given corpus.
Instead,
we follow the methodology
of \newcite{yarowsky1993} and \newcite{martinez2000},
who test the OSPC hypothesis 
by analyzing the performance of a supervised WSD system 
in which each feature corresponds to a distinct type of a collocation.
The rationale
is that
the accuracy of the WSD system 
indicates
the level of 
support for the hypothesis in the training corpus.

For the experimental evaluation,
we adopt
the IMS system of \newcite{zhong2010}.
IMS learns a separate classification model for each ambiguous word
in the training data, 
with each class corresponding to one sense of the word. 
The system employs three types of features, which broadly correspond
to different kinds of collocations:
(1)~the presence of specific content words in specific positions
relative to the focus word;
(2)~the set of POS tags in the context of the focus word;
(3)~the presence of specific content words in the {\em bag-of-words} context
of the focus word.
We train IMS on English SemCor,
and test on the concatenation of five
benchmark datasets of \newcite{raganato2017}.

The results of the experiment strongly support the OHPC hypothesis.
The test set contains 528 occurrences of words from our \hcs{}.
Six of those words,
each appearing in one instance,
are not attested at all in SemCor.
IMS selects a sense of the correct homonym in 506 out of the remaining 522 instances.
Of the 16 classification mistakes,
three 
are attributable to errors in the \nsix{},
and two are 
due to the WordNet mapping issues described in Section~\ref{data}.
Thus, the effective accuracy of IMS on the homonymous words in the test set is 97.9\%.

Analysis of the remaining 11 errors made by IMS
shows that their principal cause is insufficient training data.
For example,
the noun `match' in the sense of ``piece of wood''
occurs only once in the entire SemCor corpus,
which prevents IMS from reliably recognizing this sense.
Other obvious mistakes,
such as {\em ``follow the \underline{lead}''}
misclassified as ``metal,''
are explained by the lack of training examples 
involving the collocations that occur in the test set.
We conclude that the IMS accuracy on the test set
should be interpreted as a lower bound for the applicability of OHPC.

\subsection{One Homonym per Sense Cluster}
\label{ohpsc_exp}
   
We test 
our fourth hypothesis, OHPSC,
by searching 
an existing resource
for
clusters that contain senses from distinct homonyms.
We cannot perform this experiment on the {\nsix}
because we use it to derive our mapping from WordNet senses to homonyms
(Section~\ref{mapping}).
Instead, 
we run it 
on the high-quality, hand-crafted OntoNotes 
clustering\footnote{\it{https://catalog.ldc.upenn.edu/LDC2013T19}},
which previously used as a gold-standard
by \newcite{snow2007}.
The clustering 
includes 439 of the 1601 lemma/POS pairs
that are listed in our \hcs{}.
Those words involve 2467 WordNet senses that are grouped into 1578 clusters,
of which 1555 (98.5\%) are found to contain 
no homonymous senses,
as our hypothesis predicts.

We manually analyze
the 23 clusters that appear to combine senses from distinct homonyms.
The vast majority (21) of these apparent exceptions 
are artifacts of errors in the {\nsix}.
The errors
are easy to spot by native speakers
because senses within a single cluster 
clearly correspond to distinct 
coarse-grained senses in ODE.
In the remaining two cases, 
OntoNotes clusters two pairs of homonymous senses:
(1)~the noun `tap' as
``the sound made by a gentle blow'' and 
``a faucet for drawing water,''
and
(2)~the verb `pose' as
``introduce'' and ``be a mystery to.''
Even though we find these two clustering decisions somewhat debatable,
we treat them as actual exceptions to our hypothesis.
We conclude that 
the OHPSC hypothesis 
is corroborated in over 99.8\% of the OntoNotes clusters.

\section{Conclusion}
\label{conclusion}

We have investigated the concept of homonymy,
formulating four hypotheses that follow a common pattern.
Taken together,
our hypotheses suggest that,
figuratively speaking,
homonyms seem to repel each other,
like particles with the same electric charge.
The
experiments performed 
using our new resource
confirm that distinct homonyms
are rarely observed
in connection with
a single translation, discourse, collocation, or sense cluster.
In addition,
they demonstrate that 
contraventions of the empirical predictions made by our theory
more often than not
identify errors in existing resources.

We envisage several directions
for building upon the theoretical basis 
established in this paper.
In order to extend our {\hcs},
we plan to develop an operational method for identifying \nonetymo{} homonyms
on the basis of translation sets involving multiple languages.
We anticipate that translations extracted from parallel corpora 
will facilitate the creation of high-quality 
coarse-grained sense inventories via sense clustering.
As a step towards this goal,
we will investigate the problem of automated mapping 
between senses and translations.

\section*{Acknowledgements}

We thank 
Genna Cockburn, Amy Hua, and Jacob Skitsko
for the assistance in preparing the \hcs{}.
We thank 
Yixing Luan and Haozhou Pang
for performing additional experiments and analysis.

This research was supported by 
the Natural Sciences and Engineering Research Council of Canada, 
Alberta Innovates, and Alberta Advanced Education.

\nocite{katamba}

\bibliography{wsd}
\bibliographystyle{aaai}
\end{document}